\documentclass[acmlarge]{acmart}

\usepackage{amsmath,bm}

\usepackage{array}
\usepackage{geometry}
\usepackage{diagbox}
\usepackage{adjustbox}
\usepackage{booktabs}

\AtBeginDocument{%
  }

\setcopyright{acmlicensed}
\copyrightyear{2025}
\acmYear{2025}
\acmDOI{XXXXXXX.XXXXXXX}

\acmJournal{POMACS}
\acmVolume{37}
\acmNumber{4}
\acmArticle{111}
\acmMonth{3}

\begin{document}

\title{Foundation Models for Environmental Science: A Survey of Emerging Frontiers}

\author{Runlong Yu}
\authornote{Both authors contributed equally.}
\email{ruy59@pitt.edu}
\author{Shengyu Chen}
\authornotemark[1]
\email{shc160@pitt.edu}
\affiliation{%
  \institution{University of Pittsburgh}
  \city{Pittsburgh}
  \state{Pennsylvania}
  \country{USA}
}

\author{Yiqun Xie}
\affiliation{%
  \institution{University of Maryland}
  \city{College Park}
  \state{Maryland}
  \country{USA}}
\email{xie@umd.edu}

\author{Huaxiu Yao}
\affiliation{%
  \institution{University of North Carolina at Chapel Hill}
  \city{Chapel Hill}
  \state{North Carolina}
  \country{USA}
}
\email{huaxiu@cs.unc.edu}

\author{Jared Willard}
\affiliation{%
  \institution{Lawrence Berkeley National Laboratory}
  \city{Berkeley}
  \state{California}
  \country{USA}
}
\email{jwillard@lbl.gov}

\author{Xiaowei Jia}
\authornote{Corresponding author.}
\affiliation{%
  \institution{University of Pittsburgh}
  \city{Pittsburgh}
  \state{Pennsylvania}
  \country{USA}
}
\email{xiaowei@pitt.edu}

\renewcommand{\shortauthors}{Runlong Yu et al.}

\begin{abstract}
  Modeling environmental ecosystems is essential for effective resource management, sustainable development, and understanding complex ecological processes. However, traditional data-driven methods face challenges in  capturing inherently complex and interconnected processes and are further constrained by limited observational data in many environmental applications.
  Foundation models, which leverages large-scale pre-training and   universal representations of complex and heterogeneous data, offer transformative opportunities 
  for capturing spatiotemporal dynamics and dependencies in environmental processes, and facilitate adaptation to a broad range of applications. 
  This survey presents a comprehensive overview of foundation model applications in environmental science, highlighting advancements in common environmental use cases including forward prediction, data generation, data assimilation, downscaling, inverse modeling, model ensembling, and decision-making across domains. We also detail the process of developing these models, covering data collection, architecture design, training, tuning, and evaluation. Through discussions on these emerging methods as well as their future opportunities, we aim to promote interdisciplinary collaboration that accelerates advancements in machine learning for driving scientific discovery in addressing critical environmental challenges. 
\end{abstract}






\maketitle

\section{Introduction}


Healthy environmental ecosystems are fundamental to human survival and well-being, providing essential resources such as clean air, water, food, and energy, which are vital for sustaining life and economic development. 
Modeling environmental systems is critical for understanding the underlying processes and creating predictions to inform the management of natural resources. 
However, this is challenging because environmental systems are inherently complex, with numerous interacting processes, and are often poorly observed due to the substantial cost needed for data collection. 
Traditionally, process-based physical models have been developed for modeling ecosystems in many environmental domains, including climate science~\cite{faghmous2014big}, hydrology~\cite{xu2015data}, agriculture~\cite{jia2019bringing}, forestry~\cite{moorcroft2001method}, and geology~\cite{reichstein2019deep}. 
These methods rely on mathematical or physical equations that model relevant processes. Due to the complexity and the presence of unobservable variables, they often involve approximations or parameterizations. Calibrating these models can also be time-consuming and require extensive domain expertise.


With advances in data collection and processing in environmental science, there is growing interest in using artificial intelligence (AI) and machine learning (ML) models for modeling environmental ecosystems~\cite{bergen2019machine,hsieh2009machine,ivezic2014statistics,willard2022integrating,karpatne2017theory,karpatne2018machine,koizumi2018feedback}. These data-driven methods are particularly promising when certain processes are not fully understood or require significant computational resources. 
However, traditional ML models are typically designed for specific tasks, which limits their ability to capture the interconnectedness of various environmental processes.
For example, predicting water quality variables (such as water temperature and nutrient concentration) and water quantity variables (such as streamflow) is often handled by separate models, even though these variables are influenced by some common processes. This siloed approach hinders the ability to understand the relationships between tasks and share information effectively across models. It could also introduce model bias due to approximations that overlook certain minor processes for the target variable.  Additionally, stakeholders like resource managers, who lack expertise in AI and ML, often struggle to interpret and integrate results from multiple models to make informed decisions.


Recent ML studies have extensively explored transfer learning and meta-learning strategies~\cite{zhuang2020comprehensive,hospedales2021meta}. These works have demonstrated the potential of many ML models to be transferred from data-sufficient tasks to target data-sparse tasks.  
Building on the concept of transfer learning, there is a rising trend toward developing foundation models \cite{bommasani2021opportunities,achiam2023gpt}.  These models are pre-trained on a diverse set of pre-training tasks using either supervised or unsupervised methods to learn universal feature representations, enabling them to be fine-tuned for new tasks. Their immense success in computer vision and natural language processing~\cite{ethayarajh2019contextual,devlin2018bert,zhang2021current} has demonstrated their potential as a general framework for solving various related tasks. Such power offers new opportunities for building data-driven models to address a broad range of environmental problems~\cite{yu2025survey,xie2023geo,mai2025towards,karpatne2024knowledge}.  In particular,  many foundation models are able to harness large data from different sources and extract complex data patterns. They also offer flexibility in configuring input and output structures. For example, many large language models (LLMs) can accept user-specified inputs and generate different variables through proper prompt engineering. This is particularly useful in modeling environmental ecosystems,  where multiple related modeling tasks need to be performed, but only a subset of them is observed in each data sample.  
Additionally, these foundation models could be better adapted to new environments because they have been pre-trained on massive data.



Recognizing the transformative potential of foundation models, this survey reviews the applications of foundation models in environmental science. To cover all aspects, we gathered research from major academic databases, journals, and conferences. We reviewed leading environmental journals like \textit{Nature}, \textit{Science}, \textit{Nature Climate Change}, \textit{Water Resources Research}, \textit{IEEE Transactions on Geoscience and Remote Sensing (TGRS)}, and \textit{Remote Sensing}, as well as AI and data science journals such as \textit{Journal of Machine Learning Research (JMLR)}, \textit{IEEE
Transactions on Pattern Analysis and Machine Intelligence (TPAMI)}, and \textit{IEEE Transactions on Knowledge and Data Engineering (TKDE)}. We also explored recent developments from top conferences in AI, ML, and data mining, including the
\textit{IJCAI}, \textit{AAAI}, \textit{NeurIPS},  \textit{ICML}, \textit{ICCV}, \textit{CVPR}, \textit{ACL}, \textit{KDD}, \textit{ICDM}, \textit{SDM}, \textit{WSDM}.
Workshops like the \textit{Knowledge-Guided Machine Learning (KGML)} and the \textit{International Conference on Environmental Informatics} were considered to capture specialized research. Additionally,  preprints from platforms like \textit{arXiv} and government reports from organizations such as \textit{NASA}, \textit{USGS}, and \textit{NOAA} were reviewed to capture the latest advancements in large-scale environmental AI applications. 




The aim of this survey is to bring attention to the exciting advancements in the application of foundation models within environmental science, highlighting the opportunities for further research and development in this promising field. We hope that this survey will be valuable to both the machine learning community, by showcasing how foundation models are being developed to tackle complex environmental challenges, and to environmental scientists seeking to explore these cutting-edge models in their own work. While the focus here is on foundation models, these advancements also intersect with other domains, such as data-driven approaches in climate science, hydrology, and ecosystem management. This survey differs from existing survey papers on foundation models in that it concentrates specifically on how foundation models can adapt to environmental science, integrate environmental data from diverse sources, and offer new methods for dealing with spatial and temporal environmental processes. The paper provides a comprehensive view of how these models are being applied, developed, and optimized for environmental applications, filling a gap in the literature on the use of foundation models for scientific challenges.

We organize the paper as follows. Section~\ref{sec:overview} 
provides an overview of the evolution from process-based models to foundation models, highlighting the paradigm shift in environmental modeling.
Section~\ref{sec:objective} reviews the objectives guiding the application of foundation models in environmental science. Section~\ref{sec:method} delves into novel methods and architectures for developing these models. Section~\ref{sec:discussion} discusses opportunities for future research directions. Finally, the paper concludes by reflecting on the transformative potential of foundation models in advancing environmental science.

\section{Historical and Conceptual Overview}
\label{sec:overview}




\subsection{Environmental Modeling}

The exponential growth of diverse environmental data, like remote sensing, field measurements, and simulations, has created unprecedented opportunities to advance our understanding of environmental ecosystems. However, effectively leveraging this vast, heterogeneous data to model complex environmental ecosystems remains a significant challenge. Traditional approaches to environmental computation have evolved through several stages, each reflecting a distinct paradigm for addressing environmental challenges:

\begin{itemize}
    \item \textbf{Process-based models (1.0):} Also known as physics-based, first-principles-based, mechanistic, or theory-driven models, these models are grounded in the fundamental principles of physics, chemistry, and biology~\cite{fatichi2016overview,blocken2012ten}. Their primary goal is to analyze the fundamental mechanisms driving environmental phenomena. These models rely on differential equations and other mathematical formulations to represent key environmental processes, offering an abstract yet simplified depiction of target systems by focusing on their essential dynamics. While they are highly interpretable and effectively leverage domain knowledge, they remain approximations of reality. Furthermore, the calibration of unobserved variables or parameters can be challenging, limiting their application to complex systems characterized by high variability and sparse observational data.

    \item \textbf{Data-driven models (2.0):} 
    Data-driven environmental computation emerged as a powerful alternative, evolving from simple empirical methods~\cite{graham2008big} to machine learning-based methods, driven by the advent of big data. Unlike process-based models, data-driven approaches do not prioritize mechanistic insights but instead focus on identifying patterns, quantifying system characteristics, and predicting outcomes from observational datasets. This paradigm aligns with the ``Fourth Paradigm'' of science~\cite{hey2009fourth}, emphasizing data-intensive methodologies. By leveraging AI techniques, data-driven models excel in handling complex, high-dimensional problems with unclear boundaries or mechanisms. However, their ``black-box'' nature often limits interpretability and generalizability in data-sparse and out-of-sample scenarios. 

    \item \textbf{Hybrid physics-ML models (3.0):} To overcome the limitations of aforementioned single-paradigm approaches, process-guided or knowledge-guided machine learning integrates mechanistic insights into data-driven models~\cite{willard2022integrating,karpatne2024knowledge}. This hybrid paradigm embeds physical laws and domain knowledge into machine learning workflows to improve accuracy, generalization, and consistency with fundamental principles such as conservation laws. For instance, physics-informed neural networks (PINNs)~\cite{raissi2017physics1} incorporate differential equations into loss functions to ensure ML simulations being consistent with given physical equations. Similarly, studies in lake modeling have combined process-based components with machine learning frameworks like recurrent neural networks (RNNs) and long short-term memory networks (LSTMs), achieving better predictive performance for long-term trends by constraining predictions with ecological principles~\cite{hanson2020predicting,read2019process,jia2019physics,yu2024evolution,yu2024adaptive}.
\end{itemize}

Building on these computational paradigms, \textbf{foundation models (4.0)} represent the next leap forward. These models, which have already revolutionized domains such as natural language processing (NLP) and computer vision (CV), are set to transform environmental science by enabling holistic, scalable, and integrated modeling approaches. Foundation models excel in their ability to assimilate data from multiple sources and domains, capturing intricate patterns across diverse systems. Their success stems from pre-training on extensive and diverse datasets, which enables scalability and adaptability from well-studied systems to data-scarce or unseen environments through effective knowledge transfer~\cite{zhou2024comprehensive}.




\subsection{Emergence of Foundation Models}

Foundation models represent a significant leap in AI, characterized by large-scale pre-training on diverse datasets, which allows them to be fine-tuned for specific tasks across multiple domains. This marks a clear shift from earlier AI systems, where models were independently designed for individual tasks, requiring extensive feature engineering and domain-specific expertise~\cite{bommasani2021opportunities, zhou2024comprehensive}. Environmental science faced similar challenges, where separate models were often created for each target process (e.g., water temperature dynamics or dissolved oxygen concentration), leading to fragmented insights and a lack of holistic understanding~\cite{karpatne2024knowledge}. The limitations of traditional ML systems, particularly their inability to scale effectively across varied tasks and data sources, sparked the development of foundation models. 

The evolution of ML over the past several decades has paved the way for foundation models. The development of deep learning in the 2010s initiated a major transformation in AI, driven by the concept of representation learning. Unlike earlier models, deep learning systems could automatically extract meaningful features directly from raw data, improving their performance on complex tasks~\cite{bengio2013representation}. 
However, despite their success, early deep learning models were still task-specific, which limited their broader applicability across multiple domains. The true breakthrough came with the advent of transfer learning, which allowed models trained on one task to be adapted to another~\cite{zhuang2020comprehensive}. Transfer learning enabled the development of general-purpose models that could leverage pre-trained knowledge~\cite{weiss2016survey}. Building on this, foundation models such as BERT and GPT expanded transfer learning by training on massive datasets, allowing them to generalize across a wide range of tasks with minimal fine-tuning~\cite{kalyan2023survey,alyafeai2020survey,bao2023survey}.

A key driver of this transformation has been the rise of self-supervised learning. In the domain of language modeling, self-supervised learning enables models to learn from vast amounts of unlabeled data by predicting missing parts of the input~\cite{ericsson2022self}. The ability to learn generalizable representations in an unsupervised manner is particularly valuable in environmental science, where labeled data is often scarce or incomplete~\cite{wang2022self, ghosh2022robust, pantazis2021focus, hoffmann2023atmodist}. 
Another critical factor in the success of foundation models is their scalability. Models such as GPT-3, with its 175 billion parameters, demonstrate the impact of scaling both data and model size. This scalability allows the model to process complex input data and enables emergent capabilities like in-context learning, where the model adapts to new tasks simply by receiving natural language prompts~\cite{zhu2023chatgpt,huang2024enviroexam}. In environmental science, this capacity provides the potential to develop a unified model capable of handling multiple related tasks simultaneously~\cite{luo2023free,li2024lite}. Such models could serve as powerful tools, offering a comprehensive perspective on dynamic environmental ecosystems involving diverse processes. 
Lastly, architectural innovations, particularly the Transformer architecture introduced by Vaswani et al.~\cite{vaswani2017attention}, have been instrumental in learning complex data patterns. Transformers are adept at capturing long-range contextual information in data, which is crucial for modeling spatial-temporal processes. Their ability to handle large datasets and integrate multi-modal data (e.g., text, images, sensors) makes them invaluable for unified ecosystem modeling~\cite{xu2023multimodal,han2023survey}.

\subsection{The Current Landscape}

\begin{table}[!t]
\centering
\caption{Application-Centric Objectives and Methods for Foundation Models in Environmental Science.}
\begin{adjustbox}{max width=\textwidth}
\begin{tabular}{|>{\centering\arraybackslash}m{3.5cm}|>{\centering\arraybackslash}m{3cm}|>{\centering\arraybackslash}m{3.6cm}|>{\centering\arraybackslash}m{2.6cm}|>{\centering\arraybackslash}m{2.2cm}|>{\centering\arraybackslash}m{3cm}|}
\hline
\diagbox[width=3.5cm, height=1.5cm, innerleftsep=0.3cm, innerrightsep=0.3cm]{Objectives}{Methods} & 
4.1 Data Collection & 
4.2 Foundation Model Architecture & 
4.3 Training & 
4.4 Tuning & 
4.5 Evaluation and Diagnostics \\ \hline
3.1 Forward Prediction &\cite{stewart2024ssl4eo, nguyen2023climax, zhang2024geoscience, thulke2024climategpt, morera2024foundation, kochkov2024neural,mai2024opportunities, li2024segment, bi2022pangu,  zhang2024opportunities, bodnar2024aurora, guo2024skysense, li2024new, stevens2024bioclip, lu2024vision, ling2024improving, dong2024generative, ren2024watergpt} &\cite{li2024foundation, mai2024opportunities, thulke2024climategpt, morera2024foundation, ravirathinam2024towards, zhang2024geoscience, yu2025physics, deforce2024leveraging, hong2024spectralgpt, lu2024vision, liu2024remoteclip, chen2024rsmamba, wang2022advancing, ge2022geoscience, nguyen2023climax, bi2310oceangpt, bommasani2303ecosystem, rezayi2022agribert, lin2023geogalactica, bi2022pangu, lam2022graphcast, han2024fengwu, zhang2024opportunities, lacoste2024geo, bodnar2024aurora, guo2024skysense, li2024new, ling2024improving, dong2024generative, ren2024watergpt, stevens2024bioclip},  &\cite{li2024foundation, thulke2024climategpt, morera2024foundation, zhang2024geoscience, ravirathinam2024towards, mai2024opportunities, lu2024vision, yu2025physics, deforce2024leveraging, deng2024k2, kochkov2024neural,liu2024remoteclip, nguyen2023climax, bi2310oceangpt, morera2024foundation, bi2022pangu,  zhang2024opportunities, lacoste2024geo, guo2024skysense, li2024new, ling2024improving, dong2024generative,  ren2024watergpt, stevens2024bioclip} &\cite{li2024foundation, thulke2024climategpt, morera2024foundation, zhang2024geoscience, ravirathinam2024towards, deforce2024leveraging, lu2024vision, yu2025physics,  mai2024opportunities, sun2022ringmo, nguyen2023climax, bi2310oceangpt, luo2023free, bi2022pangu, lam2022graphcast, kochkov2024neural, zhang2024opportunities, lacoste2024geo, guo2024skysense, li2024new, ling2024improving, dong2024generative, ren2024watergpt, stevens2024bioclip} & \cite{deforce2024leveraging, thulke2024climategpt, morera2024foundation, zhang2024geoscience, mai2024opportunities, lu2024vision, yu2025physics,  kochkov2024neural, nguyen2023climax, ge2023mllm, morera2024foundation,bi2022pangu,  lacoste2024geo, li2024new, stevens2024bioclip}\\ \hline
3.2 Data Generation & \cite{stewart2024ssl4eo, thulke2024climategpt, morera2024foundation, zhang2024geoscience, mai2024opportunities, nguyen2023climax, li2024segment, lu2024vision, bommasani2023ecosystem, bi2022pangu, kochkov2024neural, zhang2024foundation, zhang2023geoscience, zhang2024opportunities, guo2024skysense, dong2024generative, stevens2024bioclip}& \cite{stewart2024ssl4eo, mai2024opportunities, thulke2024climategpt, morera2024foundation,  lu2024vision, mai2024opportunities, liu2024remoteclip, deng2024k2, khanna2023diffusionsat, kochkov2024neural, zhang2024foundation, zhang2023geoscience,  zhang2024opportunities, guo2024skysense, dong2024generative, stevens2024bioclip}&\cite{khanna2023diffusionsat, thulke2024climategpt, morera2024foundation, zhang2024geoscience, mai2024opportunities, lu2024vision, cong2022satmae, kochkov2024neural, reed2023scale, zhang2024foundation, zhang2023geoscience, zhang2024opportunities, dong2024generative, stevens2024bioclip} &\cite{zhang2024foundation, thulke2024climategpt, zhang2024geoscience, morera2024foundation, lu2024vision, zhang2023geoscience, zhang2024opportunities, dong2024generative, stevens2024bioclip} & \\ \hline
3.3 Data Assimilation &\cite{nguyen2023climax, li2024segment, bi2022pangu, stewart2024ssl4eo} & \cite{mukkavilli2023ai, nguyen2023climax, chen2024personalized, zhang2024opportunities, yu2025physics}& \cite{nguyen2023climax, yu2025physics}& \cite{li2024lite, yu2025physics}& \cite{yu2025physics}\\ \hline
3.4 Downscaling &\cite{stewart2024ssl4eo, lu2024vision, thulke2024climategpt, zhang2024geoscience, mai2024opportunities, nguyen2023climax, li2024segment, bi2022pangu, kochkov2024neural, dong2024generative, fibaek2024phileo} &\cite{tan2023promises, lu2024vision, thulke2024climategpt, zhang2024geoscience, kochkov2024neural, mai2024opportunities,   stewart2024ssl4eo, hong2024spectralgpt, nguyen2023climax,khanna2023diffusionsat,  han2024fengwu, guo2024skysense, dong2024generative, fibaek2024phileo} & \cite{sun2022ringmo, khanna2023diffusionsat, thulke2024climategpt, mai2024opportunities, kochkov2024neural,  lu2024vision, zhang2024geoscience, cong2022satmae, reed2023scale, dong2024generative, fibaek2024phileo}& \cite{khanna2023diffusionsat, thulke2024climategpt, kochkov2024neural, zhang2024geoscience, lu2024vision, mai2024opportunities, dong2024generative, fibaek2024phileo}& \\ \hline
3.5 Inverse Modeling & &\cite{gupta2024unified, ravirathinam2024combining, wang2024wavediffusion}&\cite{gupta2024unified, ravirathinam2024combining, wang2024wavediffusion}& & \\ \hline
3.6 Model Ensembling & & \cite{ge2022geoscience, rezayi2022agribert, lin2023geogalactica, kochkov2024neural, morera2024foundation}& &\cite{li2024segment} & \\ \hline
3.7 Decision-making & &\cite{nguyen2023climax, thulke2024climategpt, tan2023promises, morera2024foundation} & & & \\ \hline
\end{tabular}
\end{adjustbox}
\label{tab:Overview}
\end{table}

In summarizing the current landscape, foundation models demonstrate strong potential in addressing the complexities of environmental science, with a wide range of capabilities exemplified by the following aspects. These models excel at processing diverse and large-scale datasets, capturing intricate patterns, and improving prediction accuracy across interconnected environmental systems. By integrating multiple data modalities, they provide a holistic understanding of environmental processes that traditional methods often overlook. Their flexibility, enhanced by techniques like prompt engineering, allows them to handle varying input-output relationships and generate meaningful predictions even with incomplete or heterogeneous data. In-context learning further strengthens their adaptability by incorporating auxiliary information, such as recent observations, enabling real-time prediction refinement in scenarios with rapidly evolving conditions like extreme weather events. Furthermore, foundation models, rooted in advanced pre-training on large datasets, excel in generalizing across tasks and adapting efficiently to new challenges with minimal additional data. Unlike traditional process-based models, which often require extensive domain-specific calibration, foundation models leverage pre-trained knowledge to seamlessly adapt to a variety of scenarios, particularly in data-sparse settings. Moreover, retrieval-augmented generation enhances the utility of foundation models by incorporating real-time external knowledge, such as updated satellite imagery or the latest climate reports. This ensures that predictions remain accurate, timely, and contextually relevant while supporting the integration of scientific tools like physical simulators to improve the depth and reliability of model outputs.

Developing or selecting foundation models for environmental applications involves several key considerations, as exemplified by the following dimensions. Comprehensive data collection strategies are needed to integrate diverse sources, ensuring sufficient information to represent environmental systems effectively. Model architectures should be able to support input modalities and capture underlying spatial and temporal patterns. Model pre-training tasks need to be designed to be aligned with scientific understanding of target systems, e.g., physical laws and environmental principles, so that the learned representation can enhance prediction reliability and generalizability. Fine-tuning and prompt-tuning methods enable models to adapt to specific tasks and dynamic scenarios, ensuring their relevance across a variety of contexts. Evaluation frameworks with domain-specific metrics and diagnostics are essential for assessing accuracy, robustness, and generalizability. 

Table~\ref{tab:Overview} highlights how foundation models support application-centric objectives, including forward prediction, data generation, data assimilation, downscaling, inverse modeling, model ensembling, and decision-making, through the use of diverse methodologies. This mapping reveals the synergistic relationship between objectives and methods, providing a structured framework for understanding the potential applications of foundation models in environmental contexts. In the following sections, we delve into the specific application-centric objectives these models can achieve, alongside the opportunities they present and the challenges that should be overcome for effective implementation in environmental science.

\section{Application-Centric Objectives} 
\label{sec:objective}


Environmental ecosystems involve complex physical, meteorological, and biochemical processes that interact with each other and evolve over time. Both physical process-based and data-driven computational models have been developed to capture the underlying processes of the target system. In general, the objective of these computation models for modeling environmental processes can be expressed as $\mathcal{F}\left( \bm{x}_t, \bm{s} \right) \rightarrow \bm{y}_t$, where $\bm{x}_t$ denotes the dynamic input drivers in time $t$, $\bm{s}$ is the set of ecosystem characteristics or parameters of the systems, which is often assumed status and slowly evolving. The function $\mathcal{F}(\cdot)$ is the model producing target variables $\bm{y}_t$. 


This section examines the application of foundation models in environmental science, focusing on various application-centric objectives, such as forward prediction, data generation, data assimilation, downscaling, inverse modeling, model ensembling, and decision-making. For each objective, we begin by defining its scope, providing an overview of traditional approaches, and then illustrating how foundation models enhance these applications through their advanced capabilities. Figure~\ref{fig:application_objectives} highlights the significant advancements foundation models offer compared to traditional methods.

\begin{figure}[t]
    \centering
    \includegraphics[width=0.75\textwidth]{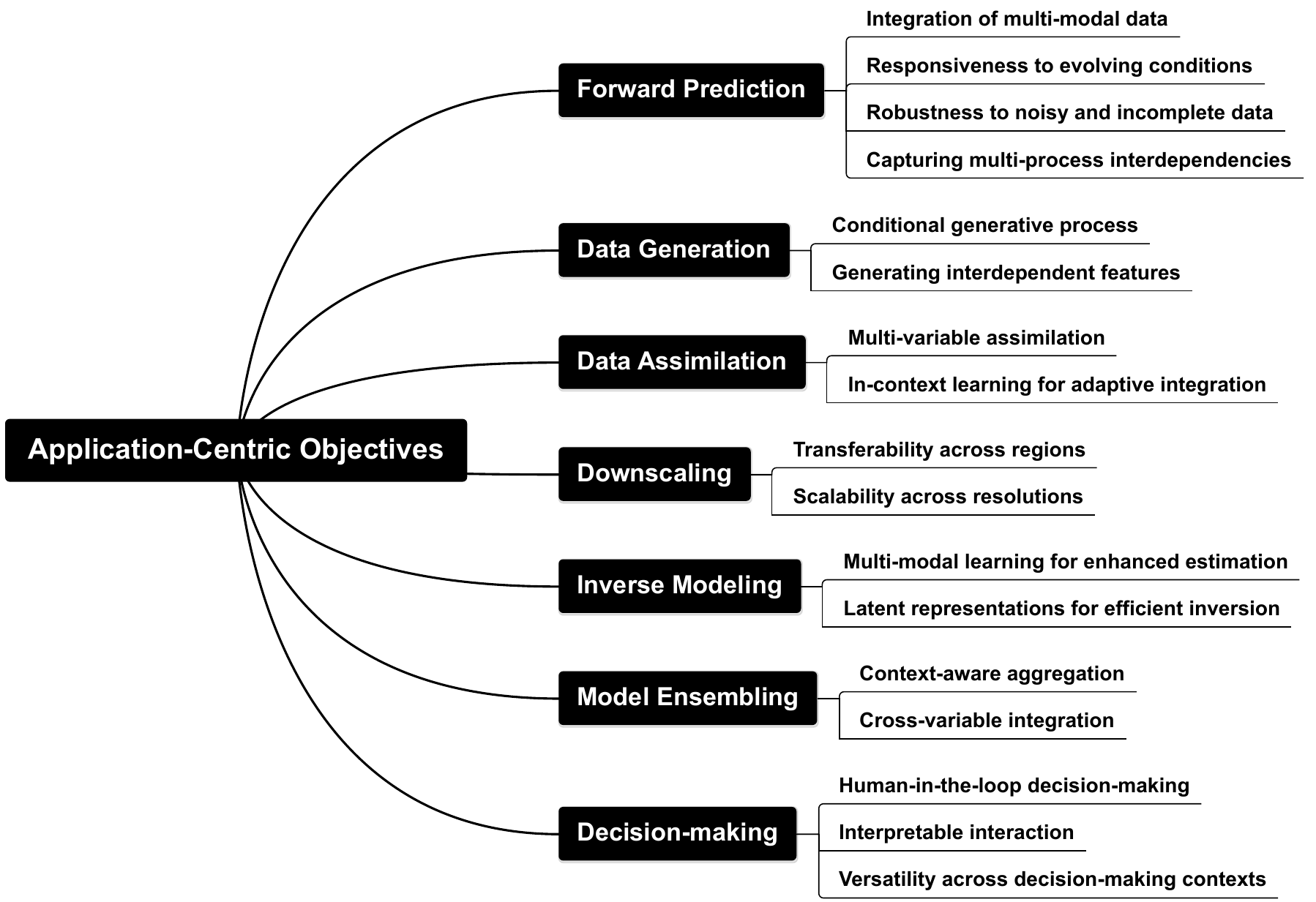} 
    \caption{Application-centric objectives and advancements enabled by foundation models.}
    \label{fig:application_objectives}
\end{figure}

\subsection{Forward Prediction}

Forward prediction is the dominant application of computational methods in environmental science, which aims at creating the model $\mathcal{F}(\cdot)$ to make predictions that align closely with observed data.
Examples include prediction of weather dynamics, carbon emission, crop yield, water quality, and pest and disease outbreaks. Standard forward prediction tasks simulate the dynamics of target variables at time $t$ given the input drivers until time $t$. This can be further extended to other formulations. For example, 
forecasting is a specialized application, which builds on prediction by utilizing historical and current data to project future states of environmental variables. This method expands the capabilities of $\mathcal{F}(\cdot)$  to forecast long-term environmental changes in the future, such as shifts in climate patterns, air and water quality, and ecological conditions. 
Anomaly detection is another important prediction task that aims to identify data points or patterns that significantly deviate from expected norms. In environmental science, this technique is used for the early detection of unusual or critical changes that may indicate underlying issues or emerging threats. It identifies anomalies that might be indicative of phenomena such as sudden changes in water quality~\cite{leigh2019framework,ni2024unsupervised,rezaiezadeh2024anomaly}, unexpected wildlife~\cite{roy2023wildect}, or irregular climate patterns~\cite{wei2023lstm}. The early detection of anomalies allows for timely intervention, which can mitigate potential negative impacts on ecosystems and human communities. For example, identifying an unexpected spike in pollutant levels in a river can prompt immediate action to address potential sources of contamination, preventing further environmental damage and protecting public health~\cite{chen2022assessment}. Similarly, detecting unusual patterns in wildlife movement can alert conservationists to possible threats to species, enabling them to take protective measures~\cite{keh2023newspanda}. 



Traditionally, process-based models have been built in many of the aforementioned applications. Due to limited understanding or excessive complexity in modeling certain processes, process-based models (e.g., for simulating phenomena in climate, weather, agriculture, and hydrology) often use parameterized processes (known as parameterization) to account for missing physics. 
For example, climate models may simplify cloud formation and interactions between land and atmosphere~\cite{brenowitz2018prognostic,gentine2018could}, 
while hydrology models often involve approximations based on soil and surficial geologic classification along with topography, land cover, and climate input~\cite{bennett2021deep}. 
Parameter calibration involves using grid searches to find parameter combinations that closely resemble observed data, enhancing model reliability. 
Building on this, reduced-order models (ROMs) distill the complete mapping $\mathcal{F}(\cdot)$ further by focusing on the most impactful variables to reduce computational demand without sacrificing essential accuracy~\cite{lassila2014model,quarteroni2014reduced}. These models allow environmental scientists to perform detailed simulations and analyses efficiently, thereby supporting faster decision-making and enabling broader scenario explorations with less computational expense.

Data-driven predictive models are increasingly used as surrogates in areas where process-based models are highly biased or computationally expensive. 
These models leverage machine learning models, like neural networks, to extract complex data patterns. 
For example, convolutional neural networks (CNNs) process spatial data, such as satellite images for weather forecasting and land cover mapping~\cite{brock2022investigating,yin2023mapping}, while RNNs and other temporal models capture temporal patterns in time series data, such as climate or pollution prediction~\cite{jung2020time,nambirajan2023climatological,liu2018td,tayal2023koopman,zhou2023deep}. In particular, the LSTM model, as a variant of RNN,  has been widely used to capture long-term data dependencies, essential for predicting long-term environmental trends. 

Multiple machine learning models can be further combined as an ensemble to improve prediction performance and robustness. This idea can be further extended to combine ML  and physics-based approaches to build hybrid predictive models~\cite{willard2022integrating}.  
Compared with traditional process-based models, ML-based surrogates provide computational advantages in forward inference,  enabling faster large-scale simulations. 

ML techniques also have the potential to handle anomaly detection tasks. For example, RNN-based models can analyze complex sequential data for time series analysis and anomaly detection~\cite{wang2022improved}. When used in modeling aquatic systems, they can detect unusual temperature patterns that might signal climate anomalies~\cite{kumar2020anomaly,ozdemir2023prediction}. Techniques like one-class SVM and isolation forest are used for high-dimensional data analysis~\cite{darrab2023anomaly,marques2023evaluation}, identifying outliers in datasets such as unusual pollutant levels in air quality monitoring~\cite{aslan2022detection,wu2018probabilistic}. The outputs of ML models can also be fed to traditional statistical methods such as Z-Score and threshold-based methods~\cite{theriault2024check,blazquez2021review,munir2018deepant}, which identify anomalies by measuring deviations from the norm. These methods could be further extended to ensure that these anomalies are physically plausible. 


Despite the promise of data-driven models, their success is contingent upon access to clean and adequate ML-ready datasets, which limits their applicability to modeling many real environmental ecosystems that require heterogeneous data sources to capture complex and diverse processes.  
In contrast to standard data-driven models, foundation models offer several opportunities in predictive modeling: (1) Existing foundation models are able to integrate large volumes of data collected from different sources (e.g., satellite imagery, sensor data,  and historical climate records), and extract patterns from such complex data.  
Some existing foundation models, such as Microsoft's ClimaX~\cite{nguyen2023climax}, IBM's climate initiatives~\cite{mukkavilli2023ai}, and NeuralGCM~\cite{kochkov2024neural},  
have shown their ability to harness diverse modalities—spatial and temporal data of different variables, satellite imagery, and text data—to enable comprehensive analysis and accurate modeling of environmental systems, crucial for reflecting the complexity of natural environments. (2) Foundation models excel in adapting to new environmental conditions (e.g., a new spatial region) without extensive model configuration. This flexibility is vital in environmental science, where dynamic factors such as climate change and urban development continuously alter ecosystems. As foundation models are pre-trained with large data, they can often be efficiently adapted to new environments with slight refinement using small training data~\cite{thulke2024climategpt, ren2024watergpt, bi2310oceangpt}. 
(3) Environmental data often contain  noise and missing values,   
arising from the  
nature of various data collection processes~\cite{gao2018review}. Existing foundation models have demonstrated encouraging results in harnessing such challenging data through data imputation~\cite{jahangiri2023wide,fang2020time} or learning by disregarding noisy and missing data~\cite{zhang2023large,li2022noise,luo2023free}. 
Such superiority helps enhance their reliability in making predictions.
(4) Environmental systems involve complex interdependencies among multiple processes and target variables. Foundation models are able to better understand and model these complex relationships by leveraging advanced deep-learning architectures. For example, Models like Prithvi can predict how land use changes affect local climate patterns by analyzing interactions between vegetation, soil types, and atmospheric conditions~\cite{Prithvi-100M-preprint,mukkavilli2023ai,yamanoshita2019ipcc}. The physics-guided foundation model predicts water temperature and dissolved oxygen in lakes, illustrating how temperature shifts affect oxygen solubility~\cite{yu2025physics}.



\subsection{Data Generation}
 

Environmental data are critically needed in many applications for making decisions and policies, advancing scientific understanding, and driving computational models. 
However, environmental science frequently encounters significant data gaps. For instance,
deploying and maintaining monitoring infrastructure is expensive, particularly in remote or inaccessible regions. Data collection during specific periods (e.g., winter time with ice cover) can also be highly challenging. This can lead to data sparsity and imbalances of data availability across space and time. 
Many process-based models require input drivers derived from remote sensing, but such data can be affected by noise in satellite imagery (e.g., clouds). 
Additionally, variables such as soil properties and land use are frequently missing in certain regions due to the high labor costs associated with field studies. 
Machine learning-based data generation methods aim to generate realistic synthetic samples 
that closely mimic real-world conditions, i.e, by approximating the distribution of driver variables $P(\bm{x}_t)$ or the joint distribution of drivers and target variables $P(\bm{x}_t, y_t)$ over space and time. 
These generated data allow scientists to analyze environmental phenomena in a controlled environment, particularly when direct experimentation or measurement is limited by environmental variability and associated high costs.

Generative machine learning models have achieved tremendous success in vision and natural language processing. These models have been at the forefront of unsupervised learning in recent years.  
The core idea behind generative models is to capture the underlying probabilistic distribution of the data in order to generate similar data. With recent advances in deep learning, new generative models such as generative adversarial network (GAN), VAE, and diffusion models have been developed. These models have shown significantly better performance in learning non-linear relationships, allowing them to extract representative latent embeddings from observation data. As a result, the data generated from these latent embeddings closely resembles the true data distribution.
In particular, GAN-based models, such as conditional GAN (cGAN) have shown great success in generating 
satellite images~\cite{kulkarni2020semantic, kim2020impact, mansourifar2023towards}, which contributes to  many important tasks, including crop yield prediction~\cite{nie2022prediction, anuradha2024enhancing,bai2024conditional} and deforestation monitoring~\cite{martinez2022comparison, boroujeni2024ic}. 
The VAE-based models have also been used to simulate artificial material samples~\cite{sardeshmukh2024material, stein2019machine} and to quantify dissolved oxygen levels in river ecosystems~\cite{stein2019machine}.




Foundation models, trained on vast and varied datasets, offer new opportunities to enhance data generation capabilities in environmental science.  
This is helpful for a wide range of environmental applications where observational data is difficult to obtain.
For example, Deng et al.~\cite{deng2024k2} utilized the first-ever LLM-based foundation model to gather and construct domain-specific data for data-limited regions to support multiple downstream tasks in geoscience research.
In particular, foundation models excel in integrating multiple types of data, such as textual, visual, and numerical formats, allowing for the accurate depiction of complex environmental conditions (e.g., through local weather, soil properties, vegetation, and optical images). Combining such heterogeneous data sources for conditioning the generative process can enhance the alignment of generated data with complex real-world conditions. 
Moreover, with the built-in attention mechanism, these models can better capture interactions among different data sources, which facilitates the generation of specific data types guided by the information of other data sources. For instance, 
Khanna et al.~\cite{khanna2023diffusionsat} proposed DiffusionSat to capture the internal relations across multiple data sources, including high-resolution remote sensing data and text-based captures, and further improve the performance in several generative tasks, including temporal generation, multi-spectral super-resolution, and inpainting.

However, concerns persist regarding these models' potential to ``hallucinate,'' meaning they may produce plausible but nonexistent or incorrect data points, especially when exposed to novel conditions not present during model pre-training. This hallucination may lead to erroneous conclusions about environmental and ecological understanding. Furthermore, the reliability of foundation models in producing scientifically actionable data is critical yet challenging, necessitating rigorous validation against established data to ensure accuracy and adherence to scientific principles, particularly in conditions where data is sparse or inherently noisy.


\subsection{Data Assimilation}

Data assimilation integrates new observational data into the computational model $\mathcal{F}(\cdot)$ to improve the accuracy of simulations and predictions. It combines diverse data sources like satellite observations and ground-based measurements with computational models, improving the precision of weather forecasts, climate models, and environmental monitoring systems. For example, in meteorology, data assimilation incorporates real-time weather observations into atmospheric models to refine predictions of weather conditions. In oceanography, it integrates measurements of sea surface temperatures, salinity, and currents to improve the accuracy of ocean circulation models.

Traditional methods, such as the Kalman filter have been fundamental in advancing data assimilation tasks in environmental science~\cite{klimova2018application}. For example, the ensemble Kalman filter (EnKF) for non-linear systems iteratively updates numerical models with real-time observational data. In meteorology, for example, the EnKF can assimilate diverse data sources like satellite imagery, radar data, and ground-based weather station observations into atmospheric models, refining initial conditions and leading to more accurate short-term weather forecasts~\cite{ott2004local}. The EnKF method is also widely used in many scientific challenges including soil moisture estimation~\cite{reichle2002hydrologic}, water properties prediction~\cite{park2020variable}, crop yield forecasting~\cite{de2007crop}, and more. 

ML models have significantly advanced data assimilation through efficient model updates in an incremental manner. In environmental science, this approach has been used to assimilate data sources like satellite observations and climate proxies, improving long-term predictions~\cite{bocquet2023surrogate, farchi2021using, yang2023flexible,zwart2023evaluating,zwart2023near}. For example, Chen et al.~\cite{chen2021heterogeneous} built a heterogeneous graph to model the river network and further introduced invertible neural networks (INNs) to continuously assimilate real-time data to adjust the model's state in long-term predictions. This approach was shown to improve the accuracy and ensure predictions are physically plausible. Following this work, Chen et al.~\cite{chen2023meta} also introduced an online learning strategy to dynamically reweight newly observed samples within a recent time window and use them to continuously refine the predictive model. 
However, standard ML-based data assimilation methods remain limited as they are mostly designed to assimilate only a specific type of variable, restricting their use when multiple types of observations are available. 
This can potentially lead to gaps in the overall data integration and effectiveness in fixing model bias over long-term modeling process.



Unlike traditional methods that typically focus on a single variable, 
foundation models offer the potential to significantly enhance data assimilation by 
assimilating diverse observational data from heterogeneous sources, such as satellites, sensors, physical simulation, and ground stations, across environmental science. This is also essential for processing varying data formats and scales, from high-resolution satellite imagery to detailed sensor data, facilitating a comprehensive and unified analysis. 
The ability to assimilate diverse data sources has been demonstrated in various environmental applications, including weather forecasting~\cite{mukkavilli2023ai, nguyen2023climax, chen2024personalized}, soil moisture estimation~\cite{deforce2024leveraging}, nitrous oxide emission, and streamflow prediction~\cite{li2024lite}. 
For instance, in aquatic systems, they can integrate temperature observations at one time and dissolved oxygen measurements at another~\cite{yu2025physics}, or both field measurements and remote sensing estimations at different frequencies, by leveraging their multi-modal capabilities to synthesize and refine predictions. 
Prior work further incorporated the knowledge about dependencies among internal processes when assimilating multiple data sources.  
For instance, in lake modeling, Yu et al.  proposed a foundation model that integrates observations of both water temperature and dissolved oxygen measurements when available~\cite{yu2025physics}. The model explicitly captures the influence of water temperature on oxygen dynamics, which enables the assimilation of  one variable to benefit the modeling of the other.


In addition, as an emergent capacity of growing parameter space, foundation models can dynamically adapt to new observations through in-context learning. Specifically, existing LLMs can incorporate recent observational examples in prompts, which allows the adjustment of model predictions to be aligned with distribution shift without extensive retraining. This capability enables foundation models 
to 
effectively respond to evolving environmental conditions and varying data availability, enhancing predictive accuracy in dynamic scenarios such as extreme weather events or seasonal changes~\cite{li2024lite}.





\subsection{Downscaling} 

Downscaling refines large-scale environmental predictions to deliver detailed local predictions by transforming coarse-resolution data from global or regional models into high-resolution outputs. This process aims to approximate a high-resolution computational model $\mathcal{F}^\mathcal{H}(\cdot)$  
that produces fine-scale output $\bm{y}^\mathcal{H}_t$. 
Downscaling is used in environmental science to bridge the gap between broad-scale models and the specific needs of fine-level information for local-scale environmental management. For example, in climate science, downscaling global climate model outputs to finer resolutions provides more precise predictions of temperature and precipitation patterns in specific regions. This detailed information is used for urban planning, agriculture, and water resource management, addressing local climate impacts, and implementing appropriate adaptation strategies. In aquatic science, 
fish seek refuge and deposit eggs in small patches within a stream reach. Downscaling to localized small stream monitoring can assist fishery managers in prioritizing efforts to protect these critical habitats. 


Traditional downscaling methods rely on either statistical techniques~\cite{gutierrez2013reassessing} or dynamic modeling~\cite{xue2014review} to refine coarse-scale environmental predictions into detailed local predictions~\cite{vonStorch1993sta, Wilby1998sta, Yousefi2019aggregate, Hamelijnck2019aggregate, Chaudhuri2020sta, Tabari2021sta, Chau2021sta}. On the other hand, ML approaches can enhance the downscaling process by leveraging vast datasets and identifying complex non-linear mapping patterns across resolutions. 
In particular, ML models have been widely used for automatically projecting the lower-resolution environmental data into higher-resolution ones, which is especially popular in the domains of climate science, hydrology, and ecology~\cite{jebeile2021understanding, yeganeh2022machine, schneider2022machine, liu2020downscaling, guevara2019downscaling, willard2022integrating}. For example, Wang et al.~\cite{wang2021deep} adopted super-resolution methods to generate higher-resolution predictions (e.g. temperature, precipitation, etc)  in different locations and times at the local scale from coarse spatial resolutions. The authors further extended this work by transferring 
the trained model from one region to downscale the precipitation in another region under a different environment. Although state-of-the-art ML methods can be used in both statistical and dynamical downscaling, several challenges remain in generating high-resolution simulations suitable for decision making. 
First, the original low-resolution data often miss important fine-scale physical patterns. As a result, the predicted fine-scale outputs can often be inconsistent with established physical laws.  
Second, in real environmental applications, downscaling often needs to be performed over different regions and at different scales, depending on the specific needs of the target application. However, existing downscaling methods are typically designed for a single target task and are unable to be generalized across regions or downscaling scales, as local variation of environmental characteristics can severely degrade performance.




Foundation models, pre-trained on extensive environmental datasets, stand a better chance of capturing broad environmental patterns. This could facilitate downscaling to specific tasks via efficient fine-tuning or prompt-based modifications. 
Such adaptability is crucial in downscaling, where requirements can significantly vary across tasks based on the needs of the target application. 
In particular, with effective pre-training, foundation models can seamlessly adapt knowledge from one region to another, allowing their application across diverse geographic and climatic contexts with minimal adjustments. For example, Dong et al.~\cite{dong2024generative} proposed SMLFR, which incorporates sparse modeling and low-frequency information with learned general patterns to enable satellite image generation across regions.
Moreover, foundation models provide opportunities to effectively handle data across various spatial and temporal scales, providing both fine-grained local insights and broader regional analysis. For example, water management decisions in different spatial regions may prioritize different objectives (e.g., maintenance of aquatic habitat, improving water quality, ensuring water supply), which require predictions at varying scales~\cite{barclay2023evaluation,barclay2024multiscale}. 
Additionally, existing foundation models such as LLMs offer flexibility to incorporate domain-specific knowledge as contextual information in the prompts. Such knowledge integration can help produce accurate downscaling predictions while simultaneously adhering to established physical rules~\cite{chen2021reconstructing,karpatne2024knowledge}.


\subsection{Inverse Modeling}

Inverse modeling refines environmental models by estimating the ecosystem characteristics or parameters $\bm{s}$ from dynamic inputs $\bm{x}_t$ and observed outputs $\bm{y}_t$. This process is used to infer unknown variables or conditions that explain the observed outcomes. For instance, prior research inversely estimate the lake characteristics (e.g., clarity and depth) using observed water temperature~\cite{tayal2022invertibility}. Hydrologists use inverse modeling to determine the sources and extent of groundwater contamination based on pollutant concentration data~\cite{ghorbanidehno2020recent}. In atmospheric science, it traces the origins of air pollution by analyzing data on pollutant dispersion patterns~\cite{wang2022city}. Similar approach is also used in geophysics to infer subsurface properties from surface observations, which helps explore natural resources and assess geological hazards~\cite{vamaraju2019unsupervised}.

Traditional calibration of physics-based models typically employs grid search or Bayesian methods to identify parameter value combinations that best align with observations or measurements. However, these approaches are often time-consuming and demand substantial domain expertise to determine appropriate variable ranges.
For example, full waveform inversion (FWI)~\cite{virieux2009overview} is an advanced technique in geoscience for creating detailed subsurface models from seismic data. It needs to repeatedly 
simulate seismic wave propagation and update model parameters to reduce the discrepancies with the true observations.  
FWI provides precise and detailed subsurface images, making it valuable for applications such as oil and gas exploration~\cite{virieux2009overview}, and earthquake seismology~\cite{fichtner2013multiscale}. However, the estimation of parameters is computationally intensive.   

To address this challenge, ML-based inverse models have been introduced as an alternative that approximates the behavior of complex environmental systems by learning from data, significantly reducing computational costs, and handling large datasets more efficiently. These approaches have been widely used in hydrology~\cite{ghorbanidehno2020recent}, photonics~\cite{pilozzi2018machine}, land surface temperature~\cite{wang2024simfair}, ecology~\cite{wang2023high}, agriculture~\cite{ravirathinam2024combining}, among many others. For example, in the field of seismic imaging, ML methods have been developed to learn the inverse mapping from observed waveform amplitudes to the velocity profile of wave propagation through various layers of the Earth's subsurface~\cite{jin2021unsupervised}. Some studies on discovering partial differential equations (PDEs) can be seen as specific examples of data-driven inverse modeling, where PDE coefficients are estimated from available simulations~\cite{raissi2018deep,rudy2017data}. Another recent approach in knowledge-guided machine learning for inverse modeling is differentiable parameter learning (dPL)~\cite{tsai2021calibration}, which uses deep learning to infer the parameters of physics-based models through gradient descent and automatic differentiation.



Foundation models can significantly enhance inverse modeling in environmental science by learning complex relationships between system characteristics $\bm{s}$ and input-output variables $\{\bm{x}_t, \bm{y}_t\}$ from vast and diverse datasets. Their key contributions to inverse modeling include:
(1) Foundation models excel at integrating heterogeneous observational data sources, such as satellite imagery, weather data, and geophysical measurements, to improve the accuracy and robustness of inverse modeling. Unlike traditional approaches that rely on domain-specific numerical solvers, foundation models leverage multi-modal learning to capture complex dependencies across diverse datasets. For instance, in environmental monitoring, combining remote sensing data with meteorological inputs can enhance crop type mapping by refining classification accuracy across varying climatic conditions~\cite{ravirathinam2024combining}. Similarly, in geophysics, fusing seismic data with geological priors can improve subsurface imaging, reducing uncertainty in parameter estimation~\cite{gupta2024unified}.
(2) Traditional inverse modeling techniques often require solving ill-posed problems through iterative numerical approximations, which can be computationally expensive and sensitive to noise. Foundation models provide a more efficient alternative by learning latent representations from observational data, which encode the underlying physical processes governing inverse relationships. This approach enables rapid and accurate inference without the need for exhaustive simulation-based optimization. For example, in seismic imaging, learning a mapping between waveforms and velocity structures within a latent space allows for faster reconstructions compared to FWI~\cite{gupta2024unified}. Likewise, in environmental science, leveraging learned feature space can facilitate the estimation of pollutant sources or climate trends from sparse observational data. By utilizing these representations, foundation models accelerate inverse modeling workflows while enhancing generalization across different environmental contexts.

\subsection{Model Ensembling} 

Model ensembling in environmental science involves integrating outputs from multiple predictive models $\{\mathcal{F}_1, \mathcal{F}_2, ..., \mathcal{F}_K\}$ to enhance the accuracy and robustness of the final output~\cite{zounemat2021ensemble, shahhosseini2020forecasting, islam2021flood, abbaszadeh2024hybrid}. This technique capitalizes on the strengths of different models through model blending, which combines diverse predictions, and post-processing, which refines these predictions based on additional observational data.
Model ensembling has been widely used across many disciplines. 
For instance, in weather forecasting, it amalgamates predictions from multiple numerical weather prediction models to create a more comprehensive forecast~\cite{gronquist2021deep, hewage2021deep}. Similarly, in hydrology, it is employed to synthesize and refine streamflow predictions from several hydrological models, ensuring more reliable and precise forecasts~\cite{troin2021generating, islam2021flood}. In climate science, post-processing adjusts model outputs with actual climate data to mitigate biases and enhance the accuracy of long-term projections~\cite{ahmed2020multi,kochkov2024neural}.


Foundation models can enhance model ensembling by offering flexibility in adjusting model aggregation and the capacity in capturing complex interactions among different processes.  
In particular, they excel in two key areas:  (1) By integrating contextual information from different sources, foundational models can facilitate the combination of multiple models in an intelligent way. 
In particular, these models can integrate diverse datasets, including atmospheric conditions, oceanographic measurements, and terrestrial data, utilizing advanced algorithms to effectively synthesize environmental conditions for each downstream task. Based on the resulting conditions,  the weight of each individual model can be adjusted accordingly.   
The contextual information in the prompts could also include prior knowledge about each individual model, such as the modeling aspects or scales each model prioritizes. This enables the selective aggregation of models most relevant to the downstream task. 
(2) By integrating the modeling of different processes, the foundation model can further extend traditional model ensembling by combining individual models of different variables in the target ecosystem. For example, the models for predicting carbon fluxes and nitrogen fluxes from agricultural ecosystems can be combined as the underlying processes depend on each other. 
Such a flexible ensemble also allows for rapid model adjustment when new observations of certain variables are available. This helps ensure ensembles remain responsive to changing conditions and specific predictive challenges.







\subsection{Decision-making}

Decision-making in environmental science applies model insights to devise strategies for natural resource management and environmental conservation, focusing on achieving sustainability goals. This requires evaluating the potential impacts of different management strategies under different simulated conditions, which in turn aids in risk and benefit assessment. For instance, in climate adaptation, decision-makers use models to predict outcomes of sea-level rise under different mitigation efforts, which are then used to guide the mitigation policies.
Predictive models are also used to 
forecast environmental risks such as floods, 
which could in turn inform proper urban planning~\cite{taromideh2022urban}. 
Based on these predicted outputs, one could design the desired objectives (often referred to as rewards) to optimize the decision-making process. 
For example, the process of resource allocation can be optimized 
to ensure sustainable use while balancing ecological and human needs, e.g., freshwater is more needed in drought-prone areas~\cite{lofton2023progress}. 
Scientific data can also be used to formulate policies to regulate 
human management practices with the aim of reducing carbon emissions or controlling deforestation~\cite{larrea2022exploring}. 
Designing such optimization objectives often requires the engagement of stakeholders to provide insights into the effect of decisions. 
For example, involving community feedback in the planning stages of conservation projects can align scientific recommendations with local values and needs, fostering more sustainable environmental management practices~\cite{zurell2022spatially}.


Foundation models in decision-making within environmental science streamline the integration of complex data into actionable strategies. 
Key advantages of foundation models include: 
(1) By directly processing the input feedback from stakeholders, foundation models may circumvent the need for intricate reward designs that are traditionally required in machine learning. 
Given such feedback, foundation models are able to update actions and policies by leveraging
comprehensive information across diverse datasets.  
This could be beneficial in many environmental decision-making contexts like resource optimization, where nuanced environmental variables are at play. 
(2) Foundation models enable meaningful interaction with stakeholders by producing human-interpretable outputs, such as visualizations that clarify predictions and recommendations. This capability facilitates transparent and collaborative decision-making processes, making the insights accessible to policymakers and community stakeholders~\cite{tan2023promises}. (3) Foundation models exhibit strong adaptability, handling diverse tasks such as linking global climate data to localized weather predictions or combining new environmental monitoring data with established models. For example, models like ClimaX~\cite{nguyen2023climax}  support a wide range of decision-making applications, from biodiversity conservation to water resource planning in dynamic conditions.

\begin{figure}[t]
    \centering
    \includegraphics[width=0.7\textwidth]{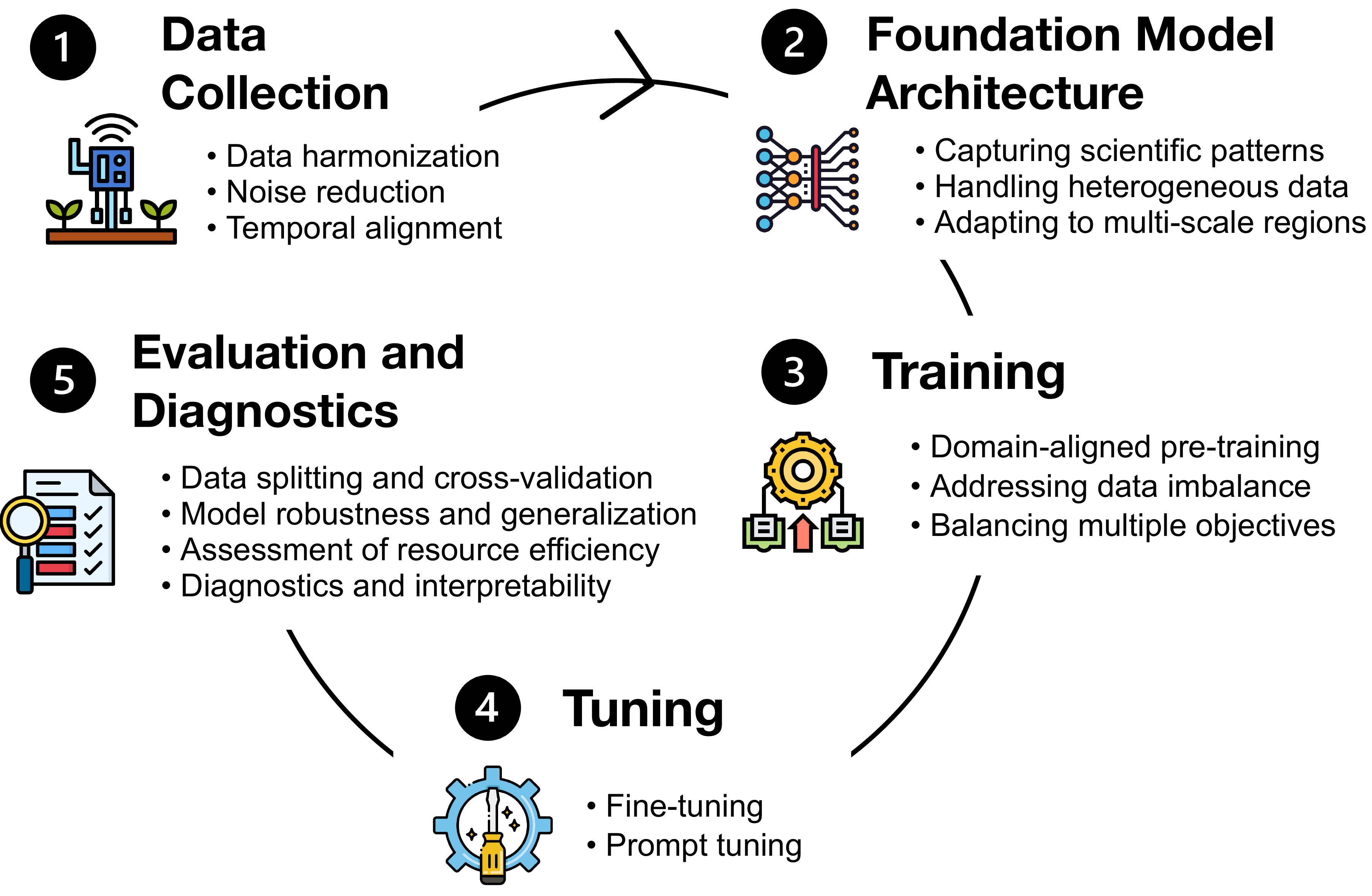} 
    \caption{Model design workflow for foundation models in environmental science.}
    \label{fig:application_objectives}
\end{figure}

\section{Method Design}
\label{sec:method}
The development of foundation models for environmental science is geared toward addressing a wide range of environmental challenges. This process initiates with the comprehensive collection of diverse data sources, such as satellite imagery, ground observations, sensor data, and historical climate records. These datasets provide the information needed for the model to recognize and interpret environmental patterns. 
Alongside data collection, selecting a robust architectural framework is essential. Options include adopting versatile LLMs that excel in handling diverse data types and intricate relationships, or constructing a customized model from scratch. 
In the training (or pre-training) stage, the model is intricately optimized to assimilate general concepts while being meticulously tailored for specific environmental tasks, with clear, measurable objectives aligned with these tasks to enhance the model’s applicability and effectiveness.
Following the initial training, the model undergoes a tuning stage that includes both fine-tuning and prompt tuning to enhance 
adaptability and scalability. These strategies are employed to boost the model's responsiveness to specific tasks or new testing scenarios. 
Finally, to fully examine the model's effectiveness and potential for deployment to real ecosystems,  an evaluation and diagnostic phase is needed to thoroughly assess the model’s performance and identify potential limitations. 

This section offers a detailed overview of the entire process involved in building a foundation model for environmental science, from data collection and model architecture to training, tuning, and evaluation. It details the specific techniques used at each step of the model design process, emphasizing how these strategies differ from conventional foundation model-building approaches and enhance the model's utility in addressing environmental science challenges.

\subsection{Data Collection}


Building robust foundation models for environmental science involves navigating several challenges inherent in environmental data collection and processing. One primary challenge is to ensure that the data encompasses a broad spectrum of distributions, which are necessary for developing models that can generalize effectively across varied environmental contexts. To achieve this, it is crucial to collect data from diverse locations, climates, and ecological systems. For example, datasets from both industry-level croplands and small-holder farmlands are needed to capture generalizable crop patterns. 
In cases where real-world data are sparse or unavailable, generating and incorporating simulated data becomes essential. Such data can be created using models that replicate environmental processes under various scenarios (e.g., different environmental conditions and management practices), allowing the foundation model to train on rare or difficult-to-capture phenomena, such as extreme weather events or long-term ecological changes.
Techniques for generating such data range from physics-based models, which rely on established physical laws, to machine learning models that learn patterns from existing data. Physics-based (or process-based) models simulate environmental processes using mathematical representations of physical laws~\cite{willard2020integrating, karpatne2024knowledge}. Examples include hydrological models like the soil and water assessment tool (SWAT)~\cite{douglas2010soil}, which models water movement in watersheds, and general circulation models (GCMs)~\cite{randall2000general}, which simulate the Earth's climate system. In addition to these, physics-guided machine learning models integrate physical laws into machine learning frameworks, enabling the generation of synthetic data that adheres to known environmental processes while capturing complex patterns not explicitly covered by traditional models~\cite{willard2020integrating, karpatne2024knowledge}. For example, a physics-guided neural network might simulate temperature variations by combining physical principles with observed temperature data.
Other techniques include agent-based models, which simulate interactions within ecosystems to assess the effects on larger environmental outcomes, such as species migration in response to climate change. Additionally, data-driven generative models, like GANs and diffusion models, generate synthetic data that mirrors the statistical properties of observed environmental data, such as creating realistic satellite images for regions lacking high-resolution imagery. These diverse approaches allow for the creation of comprehensive datasets, which are particularly valuable for variables that are difficult to observe or measure, as well as for capturing system dynamics across various scenarios.


Another significant challenge lies in ensuring that the collected data contains relevant information about the processes and variables critical to the environmental phenomena being modeled. This often requires integrating multiple datasets into a cohesive dataset, but this integration process is complex due to variations in data format, resolution, temporal frequency, and geographic coverage. For instance, satellite imagery might provide data with large spatial coverage but at a lower temporal frequency, whereas ground-based sensors might offer continuous time series data with limited spatial coverage. Furthermore, environmental data is often characterized by missing and noisy data. 
This is particularly common in remote sensing, where cloud cover can obscure satellite images, or in sensor networks where equipment malfunctions can lead to gaps in time series data. Some field measurements could also be highly localized in certain regions while being extremely sparse or completely missing for other regions. Additionally, variations in resolution can further complicate the integration of datasets.


To address these discrepancies, preprocessing steps such as data harmonization—standardizing measurement units and aligning spatial and temporal data—are essential. Noise reduction techniques, such as wavelet transforms or median filtering~\cite{boyat2013image}, are necessary to clean data from distortions introduced during collection. Moreover, careful temporal alignment is needed to address differences in temporal frequency across datasets, which ensures that the data reflects the same environmental processes over time. Merging data sources also requires careful consideration for filtering irrelevant variables while ensuring all necessary aspects of the environmental processes are represented. This might involve selecting variables like temperature, precipitation, vegetation indices, or pollutant concentrations, depending on the specific modeling objectives. Additionally, techniques like spatial weighting and data augmentation can be used to adjust the data balance in different regions, which helps address the spatial bias resulting from imbalanced or localized spatial data distribution.



Key data sources for environmental science include the National Oceanic and Atmospheric Administration (NOAA)~\cite{timofeyeva2015noaa} and the European Centre for Medium-Range Weather Forecasts (ECMWF)~\cite{ECMWF_ERA5_2021}, which provide extensive meteorological datasets pivotal for both short-term weather forecasting and long-term climate modeling. The United States Geological Survey (USGS) offers data on geological conditions, mineral resources, and natural hazards, while the Water Quality Portal (WQP) provides comprehensive water quality data across the U.S~\cite{USGS_Water_Quality_Data}. Additionally, Geographic Information Systems (GIS) and remote sensing technologies contribute crucial spatial data, with resources like the National Land Cover Database (NLCD)~\cite{homer2012national} and the HydroLAKES dataset~\cite{HydroLAKES} offering insights into land cover dynamics and water body distributions. These integrated data sources support the development of foundation models capable of addressing complex environmental challenges.

\subsection{Foundation Model Architecture}


The choice of an appropriate model architecture is driven by the specific environmental tasks and the nature of the data involved. There are two primary approaches: leveraging pre-existing models, such as LLMs or vision-based foundation models, or developing custom architectures tailored to specific environmental challenges. These challenges include handling scientific data, which often requires models capable of capturing and interpreting complex and domain-specific relationships; managing heterogeneous data (multi-modality), where data sources might vary widely in format and type, necessitating flexible and integrative models; and creating outputs for different spatial regions and modeling scales, ensuring that the models can operate effectively for downstream tasks with different task-specific data requirements. 

\subsubsection{Capturing complex patterns in scientific data}
Scientific data presents unique challenges due to its complexity, domain specificity, and often vast amounts of information that need to be processed and interpreted accurately. One effective way to handle scientific data is by leveraging pre-existing models like LLMs, which have been trained on large and diverse datasets. These models, such as GPT or BERT, are capable of understanding and generating text based on the knowledge embedded in large corpus used in their training process. In fields like climate science, hydrology, or geoscience, LLMs can be fine-tuned on domain-specific datasets to enhance their ability to interpret specialized terminology, understand complex relationships, and generate meaningful insights from the data~\cite{deng2024k2, stewart2024ssl4eo, hong2024spectralgpt, liu2024remoteclip, li2024foundation, manvi2023geollm}. For example, LLMs can be tuned towards climate science tasks using a corpus of research papers, climate models, and observational data, which helps it become proficient in understanding the nuances of climate-related language~\cite{bi2310oceangpt, luo2023free, ge2023mllm, manvi2023geollm, lacoste2021toward}. This fine-tuning allows the model to answer domain-specific questions, summarize research findings, and even assist in generating hypotheses for climate change impacts or mitigation strategies. The advantage of using pre-existing models is that they provide a robust foundation, reducing the need for extensive training from scratch while still delivering high-quality performance in specialized areas. However, while LLMs can be powerful tools, they may not always capture the intricacies of highly specific environmental data, particularly when dealing with datasets that require a deep understanding of domain-specific phenomena.

Building a customized foundation model for handling scientific data involves a strategic approach that begins with a deep understanding of the target problem and the characteristics of the data involved. Scientific data is often diverse and complex, encompassing various forms such as time-series data from sensors, spatial data from satellite imagery, and multi-modal data that integrates multiple types of information. The goal is to create a model architecture that is specifically tailored 
to capture the dynamics of the target system from corresponding datasets. Such an architecture could naturally allow the model to extract meaningful insights and make accurate predictions relevant to the scientific domain~\cite{nguyen2023climax, ge2022geoscience, rezayi2022agribert, morera2024foundation, bommasani2303ecosystem, ge2023mllm}. 

The choice of deep learning building blocks for constructing a customized foundation model in environmental science is closely tied to the nature of the data being used. For spatial data (e.g., remote sensing imagery and geospatial data),  
CNNs or graph neural networks (GNNs) are often the preferred choice. These models are adept at identifying and modeling spatial patterns, making them ideal for tasks like tracking spatial dependencies and dynamics such as water dynamics, fire propagation, and urban expansion~\cite{topp2023stream}. 
When dealing with sequential or time-series data, such as historical weather records or atmospheric $\rm{CO_2}$ levels, RNNs or transformer-based models are more suitable. RNNs are effective at modeling temporal dependencies by maintaining a memory of previous data points, which is crucial for predicting future climate trends or seasonal weather patterns. However, for 
extended sequences with long-term dependencies, transformer models, with their attention mechanisms, provide a significant advantage. They allow the model to focus on the most relevant periods, capturing long-term dependencies that are essential for understanding phenomena like climate change over decades. In addition, domain knowledge 
can be integrated into the architecture to further enhance the generalizabilty and robustness of the model. 
For example,  knowledge-guided architectures can be designed to embed physical laws (e.g., mass and energy conservation) or known cause-and-effect physical relationships into the model, ensuring that predictions are not only data-driven but also aligned with established scientific principles~\cite{liu2024knowledge,liu2022kgml, ravirathinam2024towards,yu2025physics}.
Such integration ensures that the model adheres to known scientific behaviors while still benefiting from the flexibility and learning capacity of data-driven modeling structures. 
Overall, the proper selection of deep learning building blocks can help build a reliable 
foundation model that can effectively handle the complexities of scientific data while producing insights that are grounded in domain knowledge.


\subsubsection{Handling heterogeneous data}
Existing foundation models, such as LLMs, provide the potential for 
handling heterogeneous data in environmental science problems. For example, existing models have shown the capacity in integrating multiple datasets with varying resolutions, scales, and timeframes, which is critical in many environmental science applications~\cite{barclay2024multiscale,wang2025deep,li2023point}. LLMs can also serve as a bridge between different types of data, e.g.,  dynamic data (like climate patterns) and static data (such as soil composition), by generating descriptions, summaries, or explanations of these data sources. 
Additionally, LLMs can also be used to process and analyze data of different modalities, such as textual data, such as climate reports, scientific literature, or observational logs, and extract relevant information that can be integrated with other data sources. 
Multi-modal LLMs can leverage different model components, e.g., image encoder and text encoder to handle different data modalities~\cite{tan2023promises}. 
The capability to combine text with other data sources is particularly valuable in creating comprehensive datasets where context from textual information can enhance the understanding and integration of numerical or image-based data~\cite{bi2310oceangpt, luo2023free, lin2023geogalactica, hu2023geo, li2024segment, deforce2024leveraging}. 

The ability to handle heterogeneous data becomes even more crucial for environmental science problems, as they often involve multiple processes that are observed in different datasets. Many existing foundation models can assist in pre-processing and standardizing multi-source data by generating instructions or guidelines for aligning datasets with different resolutions or scales. For example, LLMs can leverage appropriate downscaling methods for creating high-resolution climate models to match the scale of local soil data, or for temporally aligning datasets that have been collected over different periods. By leveraging the natural language processing capabilities of LLMs, scientists can more effectively manage and harmonize heterogeneous data, making it easier to build integrated models that capture the complexity of environmental systems~\cite{wang2023siamhrnet, xie2024foundation, liu2024remoteclip}. Moreover, foundation models of environmental systems could benefit from explicit modeling of cause-and-effect dependencies between different data sources. For example, the dynamics of water temperature could directly affect the concentration of oxygen and other nutrients in lakes~\cite{yu2025physics}. Incorporating such knowledge-based dependencies could be more efficient than relying on traditional methods, such as attention-based approaches or concatenation, to directly extract underlying relationships from multiple data sources.


\subsubsection{Addressing multi-region and multi-scale challenges}

Environmental processes often need to be modeled across different regions, and spatial and temporal scales, depending on the requirements of the target applications. 
Customized architectures can be created by incorporating local environmental factors, regulations, and historical data. This design allows the model to be more responsive to the unique characteristics of each region ~\cite{ghosh2023entity}. 
For instance, prior work proposed integrating static system characteristics of different locations into the model for enhancing the model generalizability~\cite{kratzert2019towards,xu2024hierarchical,chen2022physics}.  
Special architectures can also be used to generate output at different scales. For example, Fourier neural operators (FNO)~\cite{li2020fourier} and implicit neural representation~\cite{chen2021learning,chen2022videoinr,wang2023reconstruction} can efficiently make predictions at varying spatial scales by altering the output spatial grids or performing interpolation in the latent space. A major advantage of these approaches is that they require very few or no training observational data at high spatial resolutions. 
Other adaptation techniques, such as fine-tuning and prompting, can also be used for transferring the model to the target modeling scenario, which will be discussed in Section~\ref{sec:tuning}.

\subsection{Training}
In developing a foundation model for environmental science, it is essential to establish application-relevant training objectives and carefully manage trade-offs among these objectives to ensure the model's success. These training objectives act as guiding principles, ensuring the model is designed to meet the specific requirements of environmental applications. 
In this section, we will discuss these training objectives in detail.

\subsubsection{Aligning pre-training with scientific data} 
Building foundation models for environmental science requires aligning pre-training tasks with scientific data, which distinguishes them from traditional pre-training tasks like masked token prediction and instead focuses on domain-specific objectives. 
In particular, self-supervised learning has been widely used in pre-training to extract generalizable data representations from unlabeled or partially labeled  data~\cite{zhang2024geoscience, wang2022self}. 
For example, models can be pre-trained to predict critical physical variables (e.g., hydrological or atmospheric features), which guides the model to learn meaningful patterns from environmental data. 
Contrastive learning can also be used by minimizing the distance of samples with similar physical characteristics while maximizing the distance for dissimilar samples. 
This alignment with domain knowledge ensures representing data in a physically consistent latent space, which facilitates the adaptation to different downstream tasks~\cite{xie2024foundation,xu2024spatial,wan2025multi,xie2023auto}. 
Directly enforcing physical consistency can be another way to define self-supervised objectives. 
For example, physical laws such as mass and energy conservation can be directly enforced in  the learning process. 
This approach allows the model to generate predictions that adhere to scientific principles, making it more robust when dealing with unseen scenarios~\cite{willard2020integrating,karpatne2024knowledge, ravirathinam2024towards}.


\subsubsection{Addressing data imbalance}
Foundation models need to be trained with large and representative datasets. However, the real observational datasets often differ drastically in their quality and quantity over different spatial regions, which bring in significant spatial bias~\cite{xie2022fairness,he2024learning,xie2023heterogeneity}. 
Spatial fairness can be considered in the training process  to mitigate such data imbalance and facilitate the generalization of learned data patterns from data-rich regions to data-scarce areas~\cite{he2024fair,he2023physics}. 
The central idea of these approaches is to automatically adjust model behavior to achieve equitable performance across different regions and populations. 

\subsubsection{Balancing multiple objectives}
Another critical challenge is to balance trade-offs between different model priorities, such as improving prediction accuracy versus ensuring physical consistency. For instance, focusing solely on enhancing prediction accuracy can lead to models that violate physical laws, while overemphasizing physical consistency may reduce flexibility in capturing data-driven patterns from observations. 
When modeling complex systems, foundation models also need to simultaneously predict multiple interconnected variables, while also balancing the predictive performance among them ~\cite{zhu2024foundations, li2024foundation, morera2024foundation, bommasani2021opportunities}. 
Methods like multi-task learning and multi-objective optimization can be applied, allowing the model to balance competing goals effectively and share some network layers to reduce model complexity. 
By carefully navigating these trade-offs, the model can achieve a balance that satisfies both scientific accuracy and practical applicability, leading to more robust and useful foundation models for environmental science.


There are also opportunities to extend multi-task training strategies to improve performance in environmental science problems. In particular, curriculum learning strategies could be explored for the model to progressively learn knowledge from  tasks of varying complexities. For example, the training process could start from a single-variable prediction followed by multi-variable interactions. Gradient clipping techniques could also be used to stabilize the training process of large-scale models  by preventing exploding gradients from certain tasks.  The multi-task learning can also be used to simultaneously optimize model performance at multiple spatial and temporal scales, for which the modeling is often need to perform in many environmental science problems. Training on datasets spanning local, regional, and global scales can also facilitate generalization across different resolutions~\cite{li2020fourier,nguyen2023climax}.

\subsection{Tuning}
\label{sec:tuning}

Tuning is a critical phase 
for adapting foundation models to environmental science applications, during which 
the model is refined and optimized for specific tasks after the initial training phase. 
Common tuninng approaches  include model fine-tuning and prompt tuning, both of which focus on different aspects of adjusting the model to better suit specific tasks and datasets.

\subsubsection{Fine-tuning}

Fine-tuning involves taking a pre-trained foundation model and further refining it on a more task-specific dataset. In the context of environmental science, fine-tuning may require training the model on regional datasets.  
For example, if a model was originally trained on global climate data, fine-tuning could involve retraining the model on a localized dataset, such as meteorological data from a specific region, to enhance its predictive performance for regional climate forecasting. 
Alternatively, fine-tuning can use datasets focused on particular environmental downstream tasks. For example, a climate model trained globally can be fine-tuned for regional biodiversity monitoring or flood prediction~\cite{tan2023promises,wang2024gpt}.

Fine-tuning is particularly effective when transitioning models from generalized tasks (like global climate modeling) to more focused tasks (like predicting droughts in arid regions). 
Fine-tuning typically employs supervised learning techniques, where labeled datasets are used to adjust the model's parameters to improve accuracy in modeling target environmental processes~\cite{ravirathinam2024towards,luo2023free}. 
The process could also involve adjusting hyperparameters such as learning rate, batch size, and optimization algorithms to ensure the model does not overfit or underperform on the specialized dataset.
Transfer learning techniques can also be adopted in the fine-tuning process to address distribution and facilitate models trained on global datasets to be transferred to specific regions or contexts~\cite{kalanat2024spatial}. 

\subsubsection{Prompt tuning}

Prompt tuning is another technique used to adapt pre-trained large-scale foundation models towards specific downstream tasks. 
In environmental science applications, prompt tuning can provide additional contextual information to facilitate the learning of underlying data patterns for target environmental ecosystems without extensive model retraining. 
Existing LLMs can directly use this approach to guide the model’s understanding of the downstream task. Consider the species classification as an example, a prompt might include background information, e.g., ``Each biological species has a unique scientific name composed of two parts: the first for the genus and the second for the species within that genus'' before the actual classification task to clarify structure of the input and the classification task. 
One could also include other task-relevant information such as mathematical equations or physical laws that describe the relationship among key variables in the target system. 
Such a prompting method helps the model focus on the specific context of the task, improving its interpretative capabilities without altering the model's structure.

Prompt tuning can also be used in a self-augmenting manner to incorporate data semantics extracted by separate LLMs. 
For example, in an environmental monitoring task involving satellite imagery, the model might first generate a detailed language description of land features like vegetation, greenness level, water bodies, or urbanization. Then, this caption can be further incorporated into the prompt to enhance the model's response to the target prediction task (e.g., crop or forest modeling). This method can be especially useful for harmonizing multiple data modalities into LLMs. 

Zero-shot chain of thought (CoT) prompting is another powerful method that promotes reasoning by guiding the model step by step through the problem-solving process. In environmental science applications, this technique is more suitable for multi-step reasoning tasks. 
For example, after presenting a question about the effect of deforestation on local climate, the model is prompted with “Let’s think step by step,” encouraging it to generate a reasoning process that links deforestation to changes in carbon storage, local climate, and biodiversity loss. These intermediate processes could also be explicitly provided as instructions in the prompt to guide the generation process.  The second stage involves presenting the generated reasoning alongside the question to prompt the model to produce a well-informed final answer. This method is particularly valuable in environmental science, where the interplay of multiple factors often requires detailed, structured reasoning.

Prompting can also be used in other types of models, such as vision and graph-based models, to guide the adaptation to specific tasks~\cite{chowdhury2025prompt,wan2025multi}. For example, Chowdhury et al.~\cite{chowdhury2025prompt} utilized class-specific prompts in a pre-trained vision transformer to guide the model to attend to the unique image patches that reflect traits relevant for animal species classification. 
Incorporating these prompting techniques in environmental models enables more nuanced and task-specific outputs, improving the model’s adaptability without the need for extensive retraining. These methods allow models to better handle the complexity of environmental challenges 
by enhancing their understanding and processing of the input data.



\subsection{Evaluation and Diagnostics}


The fundamental goal of evaluation is to quantify how well the model performs on various tasks. Some standard metrics include classification accuracy and F1 score for classification tasks and 
root mean squared error (RMSE) and mean absolute error (MAE) for regression tasks. 
However, for environmental models, performance assessment often needs to go beyond these standard metrics used in typical machine learning settings. 
Environmental tasks are inherently multi-dimensional, requiring models to manage varying spatial and temporal resolutions, noisy data,  missing observations, and data scales. For example, the valuation of   time-series prediction of air quality or water quality needs to account for the spatial variation of target variables. Hence  R-squared is often used to measure to quantify how well the model captures data variance in each region. Moreover, 
Traditional metrics often focus on the assessment of a specific aspect of the time series, e.g., the overall magnitude match (by RMSE) or the strength of linear association between model predictions and true labels (by correlation). 
However, these metrics are unable to capture the complex nature of ecosystem dynamics and reflect 
desired ecosystem behaviors, 
such as the phase alignment of seasonal cycles, response to extreme weather events, and derivative relationships between interacting variables~\cite{collier2018international,gauch2023defense}.

Evaluation for foundation models typically involves two primary approaches: intrinsic evaluation, which focuses on the model's inherent properties and capabilities, and extrinsic evaluation, which assesses the model’s performance on task-specific datasets. 
Intrinsic evaluation involves understanding foundational properties such as robustness, generalization, or inherent biases. Since foundation models are not tied to specific tasks, this approach might assess the model’s behavior in response to broad environmental phenomena like changes in seasonal cycles or extreme weather events. For instance, measuring how well the model generalizes across diverse ecosystems (e.g., from tropical to arctic climates) would involve intrinsic evaluation, where the model’s ability to understand general environmental dynamics is tested without requiring adaptation.
Extrinsic evaluation involves adapting the model to specific environmental tasks, such as fine-tuning the model to predict climate shifts or species distribution changes. This type of evaluation, focused on the model's performance on downstream tasks, requires careful comparison across different adaptation methods. For instance, models adapted for hydrological analysis might need to be compared based on their ability to predict streamflow, with resources such as adaptation data, compute time, and required access to model architecture factored into the evaluation. 
Moreover, the evaluation of foundation models need to serve the diverse needs of different studies. In particular, different scientific communities prioritize distinct temporal patterns in their evaluations, e.g.,  agronomists emphasize growing season dynamics of crops while climatologists focus on crops' interannual variability. 

\subsubsection{Data splitting and cross-validation}

Due to the inherent spatial and temporal variability of environmental data~\cite{karpatne2016monitoring,xie2025accounting,xie2021statistically,liu2023task}, robust validation techniques are essential. K-fold cross-validation is frequently used to ensure that the model can generalize across different environmental datasets, particularly when dealing with rare events like natural disasters. 
Splitting datasets based on time (for time-series models) or geographic regions is also necessary to ensure that the model does not "leak" test data into its predictions. 
For instance, in climate prediction tasks, it is crucial to evaluate the model’s generalization across different time periods, with the aim to test its performance in both short-term forecasts and long-term projections. 
A climate model should also be tested across spatial regions, e.g., trained on European data and tested on African or Asian climates. This is particularly important for many environmental science problems, where the applicability of a model to diverse out-of-sample  conditions determines its utility. 
Leave-one-out cross-validation may be used in cases where environmental data in the target application is scarce, such as endangered species data, to make full use of limited observations.


\subsubsection{Model robustness and generalization}

Environmental data is often complex, containing missing values, noise, and bias due to localized conditions. Models should be evaluated based on how well they handle these specific characteristics. Here we discuss the following aspects: 
\begin{itemize}
    \item \textbf{Missing data:} Sensor malfunctions or gaps in satellite imagery can result in missing data, and models 
    often employ imputation approaches or 
    data-driven reconstruction to handle these gaps, or simply ignore missing values. The model 
    should be evaluated to ensure these approaches  do not introduce biases into the model's predictions and the model stays robust to missing values. 
    \item \textbf{Noisy data:} Environmental data is prone to noise, particularly in remote sensing and atmospheric datasets. Noise reduction techniques, such as wavelet transforms or median filtering, are often employed, and models should be evaluated on their robustness against noisy inputs.
    \item \textbf{Spatial and temporal resolutions:} Depending on the sensors and measuring instruments used in each region or each task, environmental data can vary significantly in both spatial and temporal granularity. 
    Foundation models need to use these data on different scales when adapted to diverse downstream tasks. 
    Models need to be evaluated for their robustness when using data of different spatial and temporal scales. 
\end{itemize}

\subsubsection{Assessment of resource efficiency }

The adaptation process for foundation models requires careful consideration of the resources involved. Evaluations should account for the data, compute time, and accessibility required to adapt foundation models for specific tasks. For example, some adaptation methods may require fine-tuning the entire model, while others may only need simple prompt adjustments. Comparing these approaches in terms of both performance and resource efficiency is crucial to selecting the most effective adaptation methods.


The scalability of foundation models should be evaluated, particularly given the large size and complexity of environmental datasets, such as global climate models or high-resolution satellite imagery. Models should be assessed for their ability to efficiently process vast amounts of data without sacrificing performance. Additionally, computational cost, including runtime and memory usage, is a key factor in determining the model's feasibility for large-scale environmental simulations.

\subsubsection{Diagnostic and interpretability}

A thorough evaluation will not only assess performance but also identify areas for improvement. 
Understanding why models make errors is essential for enhancing them in real-world applications. Error analysis could involve the identification of systematic biases, such as consistently underpredicting extreme events like floods or degraded performance for certain land cover types. Residual analysis helps check patterns in these errors, and may reveal geographic or seasonal trends that the model consistently misinterprets.

Feature importance analysis, using tools like Shapley additive explanations (SHAP)~\cite{mangalathu2020failure} or local interpretable model-agnostic explanations (LIME)~\cite{ribeiro2016local}, is valuable for understanding which variables are most influential in target applications. For example, knowing that precipitation or vegetation indices are key drivers in predicting droughts can guide future improvements in data collection or model refinement.



Finally, evaluation should account for  the strengths of each method across different aspects. For example, the comparison between 
large foundation models and traditional 
process-based climate models 
needs to consider accuracy, scalability, physical consistency, and interpretability. For example, a foundation model may outperform a traditional hydrological model in terms of predictive accuracy, but the process-based model may offer better transparency into the underlying mechanisms driving the prediction. Such insights could help users select and improve models to meet the unique demand of target applications.

\section{Discussions}
\label{sec:discussion}

In this section, we begin with a summary of the missing components in the current state of foundation models in environmental science and then explore future opportunities. 



\subsection{Summary of Challenges}




Foundation models provide innovative approaches to addressing complex environmental challenges but face unique obstacles due to the specific demands of environmental applications. In particular, beyond providing accurate ecological predictions, foundation models need to be better aligned with the goals of environmental science 
by facilitating understanding and managing complex ecosystems. To achieve this goal, current foundation models can be further improved on several aspects, including  
model interpretability, reduction of hallucination, 
uncertainty quantification, 
predicting extreme events, and supporting long-term simulation. 




\begin{itemize}
    \item \textbf{Explainability:} 
Unlike applications in industry, where the focus is often on improving accuracy and efficiency, scientific research aims to deepen understanding of the underlying processes governing complex systems. 
In environmental science, achieving this goal is often hindered by the lack of transparency in most current models. 
While some progress has been made in developing explainable AI techniques, many foundation models remain ``black boxes,'' providing limited insight into their decision-making processes. This lack of explainability restricts researchers’ ability to extract meaningful insights about environmental dynamics and to trust the model’s reasoning. For foundation models to be fully useful in environmental science, further research is needed to enhance their interpretability, allowing scientists to understand the rationale behind predictions and potentially uncover new patterns or relationships within environmental data.
    \item \textbf{Hallucination:} When foundation models produce outputs that appear plausible but are factually incorrect, it poses a significant concern in environmental science, where inaccuracies could lead to misleading conclusions and affect decision-making. Addressing hallucination remains a major challenge for current large foundation models, 
    and may require a multi-faceted approach in this field, including the incorporation of verified domain knowledge and post-processing validation. 
    Effective mitigation of hallucination is essential for developing robust, trustworthy AI systems for high-stakes applications like environmental science.
    \item \textbf{Uncertainty:} 
In environmental science, accurately characterizing uncertainty is essential for making informed decisions, especially in high-stakes areas such as climate projections, disaster forecasting, and resource management. For example, in climate modeling, decision-makers need to understand the probability and uncertainty associated with temperature rise predictions to plan effective climate adaptation and mitigation strategies. Without a clear understanding of uncertainty, policymakers may either overestimate the severity of an outcome or underestimate potential risks, leading to either inefficient resource allocation or inadequate preparedness. Uncertainty characterization is equally crucial in disaster forecasting, where foundation models predict extreme events like hurricanes, floods, or wildfires. 
If a model predicts a high likelihood of severe flooding in a particular region but fails to communicate associated uncertainties, response teams may either overreact (resulting in costly, unnecessary actions) or under-prepare (leading to unanticipated damage). Communicating uncertainty allows stakeholders to weigh risks accurately and develop contingency plans based on different scenarios. 
Despite its importance, effectively modeling and communicating uncertainty in foundation models poses several challenges. Environmental data is often incomplete or unevenly distributed, making it difficult to provide reliable probabilistic assessments, especially in regions with limited historical data or changing environmental conditions. Additionally, uncertainty estimation requires computationally intensive processes, such as Bayesian approaches or ensemble methods, which increase the model's complexity and computational demands. Foundation models should balance the need for precision with computational efficiency, especially for applications that require real-time predictions. Furthermore, communicating uncertainty is a challenge in itself. A model may produce nuanced uncertainty estimates, but conveying these in a clear, actionable format for non-expert stakeholders is complex. Misunderstanding uncertainty ranges can lead to either undue alarm or unwarranted complacency. Therefore, research is needed to develop techniques that not only quantify uncertainty accurately but also present it in an interpretable, accessible manner for diverse audiences in environmental science.

    \item \textbf{Extreme events:} 
    Predicting extreme events, such as floods, heatwaves, and wildfires, is challenging due to their rarity, and variability, and complex underlying processes. 
    Traditional training strategies, such as reweighting extreme events and data augmentation, may be insufficient for building foundation models due to the extreme  data scarcity and computational demands in certain applications. 
    Moreover, the extreme events in environmental science applications could be indicated by different data sources (e.g., remote sensing, ground-based sensors), which poses challenges for effective data harmonization. 
    Lastly, specialized evaluation metrics for quantifying the performance of extreme event detection are crucial for many high-impact real-world applications but remain largely undeveloped. 
    \item \textbf{Error accumulation over long-time prediction:} 
    Error accumulation poses a substantial challenge in long-term environmental predictions, where minor inaccuracies can compound over time, leading to significant deviations in projections~\cite{wang2023high}. This issue is especially critical in climate modeling, where small initial errors in temperature or precipitation forecasts can escalate, potentially misguiding long-term policies on climate adaptation and mitigation. Similarly, in hydrological models, error accumulation might result in unreliable predictions of river flows or reservoir levels, adversely affecting water resource management and flood risk planning. Compounding this issue, extreme events can amplify error accumulation by introducing abrupt, high-impact deviations that are difficult to model accurately. For instance, an unaccounted-for extreme event could skew baseline assumptions, exacerbating inaccuracies in subsequent projections. 
    
    \item 
    \textbf{Computational efficiency in inference:} 
    Computational efficiency in inference is critical for certain environmental applications that require high-resolution simulations, long-term prediction,  or timely responsiveness, such as disaster response. 
    Techniques like pruning, quantization, and knowledge distillation could partially mitigate this issue by removing redundant model components or reducing the model size, but they often sacrifice model perforamnce or explainability. 
    Balancing computational efficiency with accuracy, robustness, and interpretability remains a challenge, which calls for interdisciplinary collaboration across machine learning, environmental science, and computational engineering to meet the specific demands of inference-efficient models for environmental applications.
\end{itemize}



\subsection{Future Opportunities}

As foundation models continue to evolve, their application in environmental science opens up several promising opportunities for future research and development. By addressing existing challenges and expanding their capabilities, these models have the potential to transform environmental modeling, forecasting, and decision-making. Below, we outline key areas where advancements can drive significant progress. 
Future research in these areas will require fostering interdisciplinary collaboration to facilitate knowledge sharing in model design, data preparation, evaluation, and resource sharing in addressing resource constraints in computing cyberinfrastructure.



\paragraph{\textbf{Knowledge-guided machine learning:}} 
Integrating domain knowledge into foundation models, often referred to as Knowledge-Guided Machine Learning (KGML), offers a promising approach to enhance their utility and reliability in environmental science by embedding physical laws, scientific principles, or environmental constraints directly into ML-based modeling~\cite{willard2022integrating,karpatne2024knowledge,he2024knowledge,jia2021physics_tds,jia2021physics_simlr,chen2023physics,chen2022physics}. This integration ensures that outputs align with established scientific understanding while addressing the inherent complexity of environmental systems.  
One strategy is to incorporate domain-specific constraints directly into the model’s architecture or training process, embedding scientific principles and environmental constraints to guide outputs that align with known processes and reduce the risk of hallucination by creating implausible predictions. 
For instance, embedding conservation laws for energy or mass into models of aquatic ecosystems facilitate capturing the underlying heat and mass transfer processes, which enhances the prediction of water quality and quantity measures~\cite{jia2019physics,read2019process,jia2021physics_tds,yu2024adaptive}. The enforced awareness of underlying physical processes could also enhance the abilty to capture extreme events, which rarely appear in the training data.

Retrieval-augmented generation (RAG) offers another solution by allowing models to access up-to-date, verified knowledge during inference, enabling predictions grounded in current data, such as recent satellite or sensor readings. Additionally, an ontology or domain-specific database can further ensure accuracy by acting as a structured knowledge reference, allowing models to validate predictions against established environmental facts and relationships. This structured knowledge can also be integrated through dynamic retrieval, where models pull real-time information to update their predictions, or during training, where ontological constraints guide learning to reflect scientifically accepted relationships.

Post-processing validation mechanisms, including rule-based checks or a secondary model, can provide an additional layer of accuracy by filtering out hallucinations based on domain-specific standards. Human-in-the-loop approaches, with experts reviewing outputs, can further help refine model parameters and training data to reduce hallucinatory tendencies. Moreover, ontology-based checks during inference ensure that predictions maintain semantic consistency with known environmental facts, while enhanced explainability through ontological references allows researchers to trace predictions back to their factual basis, improving transparency. 

By seamlessly combining data-driven approaches with scientific knowledge, KGML bridges the gap between empirical observations and theoretical insights.  Such knowledge integration 
can help enhance the generalizability of foundation models across broader contexts with sparse data, moving beyond purely data-driven statistical relationships to reflect the interconnected processes governing natural systems, thus mitigating hallucination. 
Additionally, it can potentially 
provide actionable insights to advance the understanding of complex environmental challenges.

\paragraph{\textbf{Active learning and incremental model update: }} 
Active learning, where models identify the most informative data points for labeling, addresses a critical challenge in the application of foundation models to environmental science: the scarcity and high cost of obtaining labeled data. Environmental datasets, often fragmented and unevenly distributed across regions, limit the ability of foundation models to generalize effectively, particularly in predicting rare or extreme events like floods, heatwaves, or biodiversity loss. By leveraging active learning, foundation models can prioritize data acquisition in areas where predictions are most uncertain or where additional data would have the greatest impact. For example, a foundation model trained on global climate data might identify regions with high uncertainty in temperature projections and recommend deploying additional sensors. 
Active learning could also guide targeted field surveys in underrepresented regions to improve modeling of target environmental ecosystems. 
This approach not only enhances the quality and diversity of the data used for training but also helps address key gaps in environmental science by focusing resources on the most impactful observations.

Integrating active learning with incremental model training 
could further strengthen its utility. 
One promising strategy is periodic retraining or recalibration of model using new observational data, which 
fosters a feedback loop between model predictions and data acquisition and creates a dynamic system where models iteratively improve as new data become available. 
This can  help correct the prediction bias and reduce error accumulation over long-term predictions in real-world problems. 
Such incremental update could also be valuable in capturing extreme events, such as wildfires, by incorporating the most recent observations.  
 

Future research in this direction could focus on developing methods to seamlessly combine active learning with the multi-modal capabilities of foundation models, enabling them to suggest and adapt data collection efforts across diverse data types, from remote sensing imagery to in-situ measurements. Addressing challenges such as the computational demands of large-scale active learning and the need for real-time adaptability in dynamic environments will be essential for realizing its full potential. 

\paragraph{\textbf{Decision-making processes:}} 
Integrating foundation models into decision-making processes offers the potential to revolutionize environmental management by enabling more adaptive, data-driven strategies across a wide range of applications. For instance, foundation models can be employed to optimize decisions such as determining seeding schedules to maximize crop yield under variable climate conditions, selecting the most effective water management strategies to address both agricultural needs and urban demands, or identifying ideal locations for reforestation to enhance carbon sequestration and biodiversity. One benefit of using foundation models for decision making is in its ability to incorporate diverse data sources and consider different objectives, 
such as balancing energy production with ecosystem preservation. 
By embedding models with multi-objective optimization capabilities, decision-making frameworks can account for diverse stakeholder priorities and offer solutions that balance economic, social, and environmental goals.

Additionally, integrating decision-making into environmental simulations allows foundation models to go beyond passive predictions and actively recommend policies that adapt to changing environmental conditions. Reinforcement learning techniques, for example, enable models to learn policies through trial and error in simulated environments, refining strategies that optimize long-term outcomes. In water resource management, this could mean learning policies that dynamically adjust irrigation schedules based on real-time precipitation forecasts, while in disaster preparedness, models could simulate and recommend evacuation plans that minimize risk during extreme weather events.
This approach also holds promise for addressing global-scale challenges, such as mitigating the effects of climate change or managing shared resources across geopolitical boundaries. 
Future research could focus on enhancing the scalability and computational efficiency of integrating foundation models with decision processes, as well as addressing challenges such as uncertainty in long-term outcomes and biases in underlying datasets. These efforts will broaden the impact of foundation models, empowering policymakers and environmental managers to make more informed, adaptive decisions that address both immediate needs and long-term sustainability goals.

\paragraph{\textbf{Discover new knowledge: }} 
Foundation models hold significant promise for advancing scientific discovery in environmental science by accelerating and broadening the scope of knowledge generation  (theories, hypothesis, conjectures). These models can uncover previously unobserved patterns and relationships in large-scale, heterogeneous datasets, complementing traditional research methods. For example, they 
could propose hypotheses about the drivers of extreme weather events—such as linking urban heat island effects with localized heatwaves—or explore the relationship between deforestation patterns and biodiversity loss. 
Moreover, these models facilitate knowledge synthesis by integrating insights across disciplines, such as physics, biology, and geography, to build holistic models of complex systems. For instance, combining hydrology and atmospheric science principles can yield new perspectives on interactions between precipitation patterns and groundwater recharge rates under climate change scenarios.

Beyond discovery, foundation models can play a crucial role in hypothesis testing by integrating diverse datasets and applying computational methods to validate scientific assumptions, such as refining models of carbon sequestration by cross-referencing satellite imagery with ground-truth measurements. 
Challenges remain in many applications to ensure the outputs are interpretable, scientifically grounded, 
and will require 
deep collaboration between AI researchers and environmental scientists.


\paragraph{\textbf{Creation of opensourced benchmark datasets: } } 
The development of high-quality datasets is a cornerstone for advancing foundation models in environmental science. The reliability and accuracy of these models are inextricably linked to the caliber of the data they are trained on. By creating standardized, meticulously curated datasets that span diverse environmental processes across spatial and temporal scales, researchers can enhance model training, evaluation, and generalization.
These datasets would greatly facilitate the development of foundation models and the exploration of different algorithms in environmental science applications. 
Open-sourcing these datasets, along with pre-trained foundation models, also offers an unparalleled opportunity to catalyze collaboration across disciplines and institutions. Accessible, transparent resources enable a shared knowledge base, inviting researchers worldwide to contribute, innovate, and accelerate progress collectively. This openness democratizes the use of foundation models, reduces redundant efforts, and fosters a culture of shared responsibility for advancing environmental science.

Additionally, efforts to create and disseminate open-access datasets also call for the establishment of dedicated scientific committees to oversee their quality, representativeness, and ethical use. Such committees would play a crucial role in ensuring datasets are not only scientifically robust but also inclusive of diverse environmental contexts. Therefore, establishing standardized benchmarking protocols is crucial for assessing foundation models. These protocols should include specially designed test cases and evaluation metrics to validate model effectiveness in specific environmental applications.


\paragraph{\textbf{Reasoning based on intermediate processes: } }
    Environmental systems are governed by intricate, interconnected processes that unfold across various scales, from microbial interactions in soil to atmospheric circulation patterns affecting climate. For instance, predicting water quality in a lake requires understanding not only direct measurements (like temperature or dissolved oxygen) but also the cascade of intermediate processes that influence these measurements, such as nutrient cycles, algal growth, and seasonal temperature changes. In climate modeling, the interplay between ocean currents, atmospheric pressure systems, and vegetation cover represents a network of interdependent processes that collectively drive regional and global climate patterns. 
    Capturing these intermediate processes is contingent upon understanding and representation of cause-and-effect relationships within target ecosystems. 
    Future work could also explore 
    combining multi-source and multi-scale data to capture intermediate processes and their interactions over space and time. 
    New innovations are also needed to address a range of data issues on these intermediate processes as they can be sparse, noisy, and context-dependent. 
    Effective reasoning about intermediate processes is crucial for advancing model interpretability and mitigating hallucination in real environmental science applications.  

\bibliographystyle{ACM-Reference-Format}


\end{document}